\documentclass[runningheads]{llncs}

\usepackage[utf8]{inputenc}
\usepackage[T1]{fontenc}
\usepackage{cite} %referenser
\usepackage{amsmath}
\usepackage{wrapfig}
\usepackage{lmodern}
\usepackage{units}
\usepackage{icomma}
\usepackage{comment}
\usepackage{amssymb}
\usepackage{floatrow}
\usepackage{color}
\usepackage{caption}
\usepackage{alltt}
\usepackage{subcaption}
\usepackage{hyperref}
\usepackage{cleveref}
\usepackage{multirow}
\usepackage{graphicx}
\usepackage{bigints}
\usepackage{epstopdf}
\usepackage{bbm}
\usepackage{bbold}
\usepackage{marvosym}
\usepackage[version=3]{mhchem}
\usepackage[ruled]{algorithm2e}

\usepackage{hyperref}
\usepackage{float}
\usepackage{hhline}
\usepackage{pdflscape}
\usepackage{multirow}
\usepackage{multicol}
\usepackage{listings}
\usepackage{tikz}
\usetikzlibrary{shapes.geometric, arrows}
\tikzstyle{node} = [rectangle, rounded corners, minimum width=3cm, minimum height=1cm,text centered, draw=black, fill=gray!20]
\tikzstyle{g_node} = [circle, rounded corners, minimum width=1cm, minimum height=1cm, text centered, draw=black, fill=gray!20]
\tikzstyle{in_node} = [circle, rounded corners, minimum width=1cm, minimum height=1cm, text centered, draw=black]
\tikzstyle{arrow} = [thick, ->, >=stealth]
\tikzstyle{logical} = [rectangle, rounded corners, minimum width=3cm, minimum height=1cm,text centered, draw=black]
\tikzstyle{result} = [diamond, minimum width=3cm, minimum height=1cm,text centered, draw=black, fill=gray!20]

\usepackage{graphicx}
\title{Improved Anomaly Detection through \\ Conditional Latent Space VAE Ensembles
}
\titlerunning{Conditional Latent Space VAE Ensembles}

\author{Oskar Åström \and
Alexandros Sopasakis}
\authorrunning{O. Åström and A. Sopasakis}
\institute{Lund University}

\begin{document}
\maketitle              % typeset the header of the contribution
\begin{abstract}
We propose a novel Conditional Latent space Variational Autoencoder (CL-VAE) to perform improved pre-processing for anomaly detection on data with known inlier classes and unknown outlier classes. This proposed variational autoencoder (VAE) improves latent space separation by conditioning on information within the data. The method fits a unique prior distribution to each class in the dataset, effectively expanding the classic prior distribution for VAEs to include a Gaussian mixture model. An ensemble of these VAEs are merged in the latent spaces to form a group consensus that greatly improves the accuracy of anomaly detection across data sets. Our approach is compared against the capabilities of a typical VAE, a CNN, and a PCA, with regards AUC for anomaly detection. The proposed model shows increased accuracy in anomaly detection, achieving an AUC of 97.4\% on the MNIST dataset compared to 95.7\% for the second best model. In addition, the CL-VAE shows increased benefits from ensembling, a more interpretable latent space, and an increased ability to learn patterns in complex data with limited model sizes.

\keywords{Anomaly Detection \and Conditional VAE \and VAE Ensembles}
\end{abstract}
\section{Introduction\label{intro}}  
%Actual data is typically incomplete, faulty, missing, unbalanced and can consist of outliers. Data can have a large amount of mixed categorical and numerical features making it difficult to ascertain their relational importance. Many times there exist a size imbalance between categories. Thus a machine learning model could bias towards the larger classes in the data set. Furthermore overlap can make it difficult to identify the class each of the data points belongs to. In this case, anomaly detection can be challenging since an anomalous data point for one class may be a normal point for another. Last but not least data could be incorrectly labeled which in turn can lead to severe identification and classification or clustering errors. Finding data which is somehow anomalous to the intended distribution is therefore a crucial component of any process working with incoming real-world data. As such, anomaly detection is one of the most important problems in manufacturing \cite{Mart}, cyber security \cite{Schubert}, medical imaging \cite{Zenati}, fraud detection and several other fields. At the same time detecting anomalies is a difficult, heavily data dependent problem which relies on both data quantity and quality. The problem becomes even more complicated when considering an \textit{unsupervised} setting such as we do here. 

In real-world scenarios, data often presents itself as incomplete, corrupted, missing, imbalanced, and potentially inclusive of outliers. This data complexity, along with the presence of numerous mixed categorical and numerical attributes, poses significant challenges in determining the connections between attributes. A prevalent issue is the disparity in class sizes within datasets, leading to a potential bias in machine learning models towards more dominantly represented classes. Additionally, the presence of overlapping data points across classes complicates the task of accurate classification or clustering, rendering anomaly detection particularly arduous since what may be considered an anomaly in one class could easily be a typical instance in another. Moreover, the risk of erroneous labeling further increases the difficulty in reliably identifying and classifying data points, making the detection of anomalies that deviate from expected behaviour a critical task in any data-driven process. Consequently, anomaly detection emerges as an important concern in many domains, including manufacturing \cite{Mart}, cyber-security \cite{Schubert}, medical imaging \cite{Zenati}, and fraud detection \cite{hilal2022financial}, among others. Yet, identifying anomalies poses complex challenges that hinges on both the volume and quality of data. This challenge is intensified in an \textit{unsupervised} context, as explored in this study, where the absence of outlier-inlier labels increase the task difficulty. 

%There are many techniques to detect anomalies, each depending on the type of data available and the particular application \cite{review}. In this paper, we define anomalies to be data points which occur in low probability regions of the dataset. This will become mathematically precise in Section \ref{sec:metrics}. We are proposing a novel Conditional Latent space Variational Autonencoder (CL-VAE), which takes advantage of known normal classes in training to model normality clusters in the low-dimensional latent space using a Gaussian mixture model (GMM). The degree of anomaly is evaluated using the probability of not belonging to one of these latent clusters. More importantly, however, our proposed framework will be able to work with a broad category of data in an unsupervised manner. This novel network is then compared to established methods of anomaly detection.
Various methods exist for anomaly detection, tailored to the specific nature of the data in question and its intended application \cite{review}. Within the scope of this research, we categorize anomalies as data points that reside within regions of the dataset characterized by low probability, a concept that we explain in more detail in Section \ref{sec:metrics}. We introduce a novel approach through the Conditional Latent Space Variational Autoencoder (CL-VAE), leveraging the knowledge of normal data classes during the training phase. This approach allows for the modeling of normative data clusters within a condensed latent space by employing a Gaussian Mixture Model (GMM). The degree of anomaly is evaluated using the probability of not belonging to one of these latent clusters. Notably, our methodology extends its applicability to a wide array of data types in an unsupervised manner, setting a new benchmark by juxtaposing our network with traditional anomaly detection techniques.

We consider data which is unmarked in terms of containing anomalies or not. Nevertheless each data point is still labeled as belonging to a specific class. These class labels, which the model is conditioned upon, are only used in training and are not needed during evaluation, thus enabling online implementation into real-time systems. Such an implementation reflects real world situations where the model encounters data of many known types, but where it is of interest to identify if unknown data types are observed. This approach can furthermore be used for a wide variety of problems, flagging of new species in wildlife camera traps, or separation of goods and materials in recycling systems.

The proposed approach delves into the \textit{latent space} of the Variational Autoencoder (VAE) \cite{VAE} for deeper comprehension and analysis. The latent space, characterized by its reduced dimensionality, forces the data to approximate a certain prior distribution. We will demonstrate that the designed latent space not only preserves but also reveals the relational information concealed within the data, thereby enabling a comparative analysis of data features. We call this model the \textit{Conditional Latent Space Variational Autoencoder} (CL-VAE), and note that it builds upon previous work \cite{norlander2019latent} by the authors. 

\noindent The proposed CL-VAE incorporates three novel techniques:
\begin{itemize}
    \item \textbf{Multiple Latent Gaussians:} 
    By departing from the conventional single Gaussian assumption in the latent space, we allow for distinct distributions for different classes, thereby decreasing class overlap and confusion.
    \item \textbf{Radial Latent Space Separation:} By introducing fixed cluster centers along the circumference of a circle, we force an empty space in the center of the latent space where anomalous points tend can congregate. This radial separation is crucial in order to separate anomalous and normal points.
    \item \textbf{Latent Space Ensembles:} By using an ensemble of encoders, multiple latent spaces are formed. These spaces are merged to a group consensus space where anomalous points are distinguished to a greater extent.
\end{itemize}

%This work  proposes a methodology for class-conditional anomaly detection by augmenting the VAE's latent space with a Gaussian mixture model. 
We begin in Section \ref{sec:background} with a succinct literature review, establishing the novel aspects of our approach compared to existing methodologies. Subsequent sections delve into the theoretical underpinnings of VAEs, particularly focusing on the conditioning of our VAE to accommodate multiple Gaussians and to categorize data within the latent space effectively. The formulation of the loss function, pivotal for cluster formation, is discussed in Section \ref{sec:cvae}, where we delineate specific reconstruction and regularization terms. The paper's innovative contributions are elaborated in Sections \ref{sec:radialseparation}-\ref{sec:ensembles}, followed by a description of the datasets and experimental setup in Section \ref{sec:datasets}. Anomaly detection outcomes are presented in Section \ref{sec:results}, and the manuscript concludes with a discussion and a summary in Section \ref{discussion}.

\section{Background and State of the Art\label{sec:background}}

%There are a number of approaches to anomaly detection using generative models; many, using the reconstruction error \cite{bologna} \cite{NUS} or the loss \cite{oskar_anomCVAE} as a measure of "degree" of abnormality . However in this paper we implement a VAE as a pre-processing algorithm in order to construct a GMM latent space, where the anomaly detection occurs directly. In order to do so effectively, we introduce the CL-VAE which shapes the latent space by discovering which distribution each of the data belongs to, thus making it easier to work with. 

%As mentioned earlier, Variational Autoencoders (VAEs) are generative models introduced in 2014 by Kingma and Welling \cite{VAE}. The purpose of such a model is probability density estimation. Meaning that a VAE can estimate any underlying distribution in a dataset and start to generate new samples from that dataset. 

Various methodologies utilize generative models for anomaly detection, with several leveraging reconstruction error \cite{bologna, NUS} or model loss \cite{oskar_anomCVAE} to quantify the extent of abnormality. This study adopts a Variational Autoencoder (VAE) as a preliminary algorithm to devise a Gaussian Mixture Model (GMM) within the latent space, where anomaly detection is directly performed. To achieve this, we introduce the Conditional Latent Space Variational Autoencoder (CL-VAE), which effectively structures the latent space by identifying the specific distribution of each data point, thereby simplifying the anomaly detection process.

Variational Autoencoders (VAEs), introduced by Kingma and Welling in 2014 \cite{VAE}, are generative models aimed at probability density estimation. The generative capabilities of the VAEs allows for the generation of new data samples using the distribution of the original dataset.

Dimensionality reduction, a critical aspect of VAEs, allows for effective data compression, particularly notable in the latent space. VAEs, through this process, allow a mapping from a lower-dimensional space to a higher-dimensional one, at the inevitable cost of information loss. This information loss is measured using the reconstruction loss, which represents the negative log-likelihood of $p_{\theta}(\hat{x}|z)$, serving as a metric to minimize during the training phase. This incentivizes improvements in the encoder's efficiency for distilling essential features into $z$ from the input data $x$. Moreover, VAEs' versatility extends to processing both categorical data and facilitating non-linear transformations, distinguishing them from methodologies like PCA \cite{PCA}.

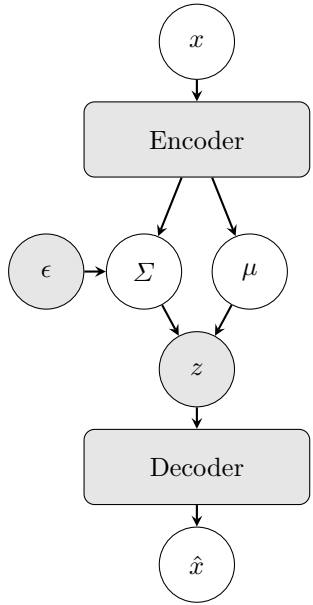
\begin{wrapfigure}[28]{r}{0.41\textwidth}
	\centering
	\begin{tikzpicture}[node distance=1.3cm]
	\node (output) [in_node] {$\hat{x}$};
	\node (decoder) [node, above of=output] {Decoder};
	\node (z) [g_node, above of=decoder,] {$z$};
	\node (mu) [in_node, above of=z, xshift=20] {$\mu$};
	\node (sigma) [in_node, above of=z, xshift=-20] {$\Sigma$};
	
 \node (epsilon) [g_node, left of=sigma] {$\epsilon$};
	\node (encoder) [node, above of=z, yshift=50] {Encoder};
	\node (input) [in_node, above of=encoder] {$x$};
	\draw [arrow] (input) -- (encoder);
	\draw [arrow] (encoder) -- (mu);
	\draw [arrow] (encoder) -- (sigma);
	\draw [arrow] (epsilon) -- (sigma);
	\draw [arrow] (mu) -- (z);
	\draw [arrow] (sigma) -- (z);
	\draw [arrow] (z) -- (decoder);
	\draw [arrow] (decoder) -- (output);
	\end{tikzpicture}
	\caption{Schematic of a typical VAE. The encoder neural network maps the input $x$ to the latent space described by the Gaussian distribution $z$. Then the decoder neural network creates a reconstruction of the input in $\hat{x}$.}
	\label{fig:VAE}
\end{wrapfigure}

A VAE architecture incorporates an encoder and a decoder neural network, surrounding a stochastic latent layer, as depicted in Figure \ref{fig:VAE}. The encoder, denoted by the posterior $p_{\theta}(z|x)$, analyses the input $x$ and generates parameters for a Gaussian distribution symbolized by $z$ in the latent layer, as also indicated in Figure \ref{fig:VAE}. The decoder, represented as $p_{\theta}(\hat{x}|z)$, uses the Gaussian distribution $z$ from the latent space to approximate the parameters describing the probability distribution of the original data. The algorithms structural composition, including the neural networks' weights and biases, is encapsulated by $\theta$.

Significantly, the latent space in a VAE is populated by estimating the mean and variance of the points, denoted by $\mu$ and $\Sigma$, respectively. This is accomplished by the mapping of the stochastic layer, characterized by $\epsilon$, $\mu$, and $\Sigma$, ensuring that each point in the latent space possesses actionable attributes due to the variance assigned to it. This unique feature of VAEs, unlike traditional autoencoders, facilitates the retention of relational information among data points within the latent space, thus enhancing their analytical relevance \cite{Chollet}.

In contrast, Generative Adversarial Networks (GANs) \cite{GAN}, address similar generative challenges employing a game-theoretic approach. However, VAEs, grounded in a robust statistical framework and their subsidiary generative capability, have been selected for our exploration due to their statistical interpretability and the simplicity with which they can be manipulated \cite{TrainVAE}.

\subsection{Conditioning and VAEs}
The incorporation of a Gaussian Mixture Model (GMM) as a training prior within a Variational Autoencoder (VAE) framework is not novel, with precedents established by \cite{RuiShu, GMVAE, GMMVAE}. Distinctly, the GMVAE approach \cite{GMVAE} diverges by not conditioning on pre-determined labels, opting instead to deduce the prior distribution via Monte Carlo methods. This concept is similarly explored by J. Su \cite{one-step}, albeit with an unnecessary classifier for our methodology, given our direct assignment of classes to specific Gaussians during training without recourse to class labels during testing. This methodology contrasts with the DEC \cite{DEC} and VaDE \cite{VaDE} algorithms, which incorporate unsupervised cluster assignments, a strategy our algorithm avoids.

Differing fundamentally from the Conditional Variational Autoencoder (CVAE) models cited in \cite{CVAE, oskar_anomCVAE, Tut}, our approach conditions the latent space directly on class labels by selecting the appropriate Gaussian. This contrasts with the CVAE's approach of maintaining a singular Gaussian assumption within the latent space while conditioning both the encoder and decoder. Furthermore, the CVAE concatenates the class label either to the encoder input or in the latent space, meaning that class labels are required in testing. Our model, on the other hand, only incorporates the class labels in the loss function, thereby removing the need for labels in evaluation. Our method also deviates from assumptions of latent and input variable statistical independence \cite{KMRW, CVAE}, and introduces unique techniques such as manual cluster separation in the latent space and the utilization of a novel latent space ensemble merging through multiple CL-VAE model iterations. These innovations are detailed in section \ref{sec:method}.

Anomaly detection in our model diverges from traditional approaches that utilize reconstruction error or loss as anomaly indicators \cite{bologna, NUS, oskar_anomCVAE}, favoring instead a classification based on latent space Gaussian representations. This method leverages the model's conditional properties and class-specific Gaussians for precise anomaly scoring, detailed further in section \ref{sec:metrics}.

The empirical validation of our methodology uses three diverse datasets: MNIST \cite{MNIST}, Fashion-MNIST \cite{FashionMNIST}, and CIFAR-10 \cite{Cifar10}, chosen for their simplicity and prevalence in machine learning benchmarks. Most anomaly detection benchmarks lack class labels for non-anomalous data \cite{ADBench, NumentaBench} and can therefore not be used for conditional anomaly detection. Therefore, we use these common datasets for our comparisons, which we elaborate further on in section \ref{sec:datasets}.

Our evaluation encompasses training both the traditional VAE and our CL-VAE model to generate distinct latent spaces, followed by an anomaly detection analysis. We opt for 2 and 3-dimensional latent spaces for visualization purposes, although we acknowledge potential performance gains from higher-dimensional representations. 

%%Benchmarking includes comparisons with standard clustering algorithms, $k$-means \cite{k-means}, and EM \cite{em}, alongside anomaly detection techniques such as Isolation Forest \cite{isolation_forest}, Local Outlier Factor (LOF) \cite{LOF}, and one-class Support Vector Machines (OCSVM) \cite{OCSVM}, further discussed in section \ref{sec:evaluationcomparison}.

\section{Theoretical Framework for the CL-VAE} \label{sec:method}

In our exploration of the Conditional Latent Space Variational Autoencoder (CL-VAE), we establish a solid theoretical groundwork essential for its development, starting with an insightful overview of the loss function within variational autoencoders (VAEs). The selection of a loss function plays a pivotal role in the neural network's performance, where an improper choice can lead to the model's failure in addressing the task at hand. The primary challenge lies in computing $\log p_{\theta}(\hat{x}|z)$ due to the intractability of the "evidence" $p_{\theta}(x)$, as elaborated in Appendix A in the supplementary material.

To navigate this complexity, we advocate for optimizing a lower bound on the likelihood, thereby sidestepping the direct computation of the intractable evidence. This approach, detailed in Section \ref{sec:cvae}, enables the effective training of the VAE by maximizing an estimable objective.

Moreover, the CL-VAE introduces several novel methodologies to refine the functionality and structure of the latent space:
\begin{itemize}
    \item The concept of \textit{radial latent space separation}, discussed in Section \ref{sec:radialseparation}, presents a new technique for organizing the latent space to better differentiate clusters and improve anomaly detection.
    \item \textit{Latent space ensemble merging}, introduced in Section \ref{sec:ensembles}, leverages multiple latent representations, merging them into a consensus space that enhances the model's predictive accuracy and robustness.
\end{itemize}

This paper also lays the mathematical foundation for understanding VAEs, setting the stage for the CL-VAE's innovative contributions. The mathematical exposition paves the way for the detailed discussion on novel CL-VAE methodologies, effectively merging theory with practical application in enhancing variational autoencoder technology.

    \label{eq:log5}
%\end{align}
%The first term above is possible to compute directly from the decoder and sampling with the reparameterization trick \cite{VAE}, which will be discussed later. The second term is a KL-divergence between a general and a normal Gaussian which has a closed form solution based on Property \ref{prop:KL_gauss} of Appendix B. Finally, the problematic last term can be shown to be strictly positive through Property \ref{prop:KL} in Appendix B. 
%Therefore we can define
%The lower bound of \eqref{eq:log5} is shown \cite{KMRW, VAE} to be,
% ELBO is a lower bound and not a pure optimization because of the final term in \eqref{eq:log5}. 
%\begin{align}
    %\begin{split}
%        L(\theta, \phi; x) = -\mathbb{E}_{z\sim q_\phi (z|x)}\Big[\log p_\theta(x|z) \Big] +  D_{KL}(q_\phi(z|x) || p_{\theta}(z)).
%        \label{eq:elbo}
    %\end{split}{}
%\end{align}
%The task is to minimize the above with respect to the parameters $\theta$ and $\phi$,
%\begin{equation}
%\phi^*, \theta^* = argmin_{\phi, \theta} L(\theta, \phi; x).
%\label{eq:phi_theta_estimate}
%\end{equation}
%The expected value in 
%The only remaining issue is to resolve the expected value in 
%\eqref{eq:elbo} can be resolved with a reparameterization idea from 
%\cite{VAE}. We outline this in the Appendix D for completeness.
\subsection{Loss Function of Variational Autoencoders}

Variational autoencoders (VAEs) establish a probabilistic framework for learning latent representations of data, where the loss function plays a crucial role. Beginning with the assumption that latent variables $z$ can be inferred from the data $x$ through the distribution $q_\phi(z|x)$, we focus on estimating the log likelihood $\log p_{\theta}(x)$. The derivation is initiated with the expectation over the latent variables, leading to a crucial expression:
\begin{align}
    \log p_\theta(x) &= \mathbb{E}_{z \sim q_\phi (z|x)}\Big[\log p_\theta(x|z) \Big] - D_{KL}(q_\phi(z|x) || p_{\theta}(z)) + D_{KL}(q_\phi(z|x) || p_\theta(z|x)), \label{eq:log5}
\end{align}
where the Kullback-Leibler (KL) divergence measures the discrepancy between two probability distributions, offering an estimate of the "distance" between them. This metric's utility is further discussed in Appendix B in the supplementary material and cited works \cite{SK}.

The log likelihood of data under the VAE model is reframed in terms of a lower bound, known as the Evidence Lower BOund (ELBO), where last term of (\ref{eq:log5}) can be shown to always be strictly positive (Appendix B). To maximize the log likelihood of data under the VAE model, we use the negative ELBO as our objective function under training:
\begin{equation}
    L(\theta, \phi; x) = -\mathbb{E}_{z\sim q_\phi (z|x)}\Big[\log p_\theta(x|z) \Big] +  D_{KL}(q_\phi(z|x) || p_{\theta}(z)). \label{eq:elbo}
\end{equation}
Minimization of this function with respect to parameters $\theta$ and $\phi$ leads to the optimal model parameters, $\phi^*, \theta^*$, which best approximate the data distribution:
\begin{equation}
\phi^*, \theta^* = \arg \min_{\phi, \theta} L(\theta, \phi; x). \label{eq:phi_theta_estimate}
\end{equation}
The expected value term in the ELBO is tractably approximated using the reparameterization trick, detailed in Appendix D, facilitating the optimization process \cite{VAE}.

This foundational section delineates the theoretical underpinnings essential for the CL-VAE's operation, elucidating how the VAE's loss function integrates with its architecture to model data distributions effectively.

\newpage
%\subsection{The change of variables idea}
\subsection{Conditioning the Autoencoder\label{sec:cvae}}

Instead of relying on a single Gaussian in latent space to approximate the unknown probability distribution \(p_{\theta}(x)\), as is the case with Variational Autoencoders (VAEs), our methodology enhances the representation of the dataset \(x\) by employing multiple Gaussians. This approach takes advantage of the inherent information present in the data, specifying these Gaussians by conditioning on any selected label from the dataset. For instance, in the context of the MNIST dataset of images, conditioning is performed based on the numeric label associated with each image.\\

%VAEs are designed to estimate the unknown probability distribution $p_\theta(x)$ for the given dataset $x$ using a Gaussian in latent space. Instead we propose to use  information, which is already available in the data, in order to more accurately express the dataset $x$ through several Gaussians. We specify these Gaussians in latent space by conditioning on any one of the labels in the data set. For the MNIST data set of images for instance we condition on the number label associated with each image. 

Following ideas for the standard autoencoder loss in  (\ref{eq:elbo}), we now derive the loss function for CL-VAE, conditioned on the class $y$, 
\begin{align}
\begin{split}
    L(\theta, \phi; x) &= L_{rec} + L_{KL} =\\
    &=- \mathbb{E}_{z\sim q_\phi (z|x)}\Big[\log p_\theta(x|z) \Big] + D_{KL}(q_\phi(z|x,y) || p_{\theta}(z|y)).
\end{split}
\label{eq:elbo2}
\end{align}

The first term in the equation above is traditionally referred to as the reconstruction term since it is a measure of the likelihood of the input image given the latent representation. Practically, this is approximated through sampling, which in practice translates to the mean squared error of the reconstruction \cite{VAE}. The second term in (\ref{eq:elbo2}) can be further expressed as,
%Traditionally, the equation's initial term is known as the reconstruction term, reflecting its role in gauging the input image's likelihood based on the latent representation. This likelihood is typically estimated by sampling methods, which effectively equate to computing the reconstruction's mean squared error \cite{VAE}. Furthermore, the expression for the second term, as outlined in (\ref{eq:elbo2}), can be expanded to provide additional clarity,
\begin{equation}
\begin{split}
    L_{KL} = D_{KL}(q_\phi(z|x,y) || p_{\theta}(z|y))= \mathbb{E}_{q_\phi(z|x,y)}\Bigg[\log\frac{q_\phi(z|x,y)}{p_\theta(z|y)}\Bigg].
\label{eq:KL_log}
\end{split}{}
\end{equation}
%To produce a closed form solution for $D_{KL}(q_\phi(z|x,y) || p_{\theta}(z|y))$ in the non-normal Gaussian case we follow ideas from \cite{one-step} and let $q_\phi(z|x)$ be some Gaussian distribution that is conditioned on $x$, and further that $p_\theta(z|y)$ be Gaussian with mean $\mu_y$ and variance 1,

To derive a closed-form solution for \(D_{KL}(q_\phi(z|x,y) \| p_{\theta}(z|y))\) in contexts diverging from the normal Gaussian framework, we utilize insights from \cite{one-step}. In this approach, \(q_\phi(z|x)\) is conceptualized as a Gaussian distribution that is conditioned on \(x\), and \(p_\theta(z|y)\), in turn, is modeled as a Gaussian with mean \(\mu_y\) and a variance of 1. This is where our model differentiates itself from traditional VAE, which assume a mean of 0 instead of \(\mu_y\).
 \begin{equation}
    q_\phi(z|x) = \frac{\exp\Big\{-\frac{1}{2} \Big|\Big|\frac{z-\mu(x)}{\sigma(x)}\Big|\Big|^2 \Big\}}{\prod_{i=1}^d \sqrt{2\pi\sigma_i^2(x)}}, \hspace{0.5cm} p_\theta(z|y) = \frac{1}{(2\pi)^{d/2}} \exp\Big\{-\frac{1}{2}||z-\mu_y||^2 \Big\}.
\label{eq:latentgaussian}
\end{equation}
This is used to simplify the $L_{KL}$ term to
$$
L_{KL} =
-\frac{1}{2}\left[d - ||\sigma(x)||^2 - ||\mu(x)-\mu_y||^2+ \sum_{i=1}^d \log\sigma_i^2(x)\right].
$$
We provide the proof for the above in Appendix C in the supplementary material. 
Furthermore, we introduce a convex combination between the two loss terms using the hyperparameter $\alpha$ which we found to be optimal at $\alpha=1/6$,
\begin{equation}
    \begin{split}
        L(\theta, \phi; x) = \alpha L_{rec} + (1-\alpha) L_{KL}.
    \label{eq:loss_function}
    \end{split}{}
\end{equation}

\subsection{Radial Latent Space Separation}\label{sec:radialseparation}

The latent Gaussian distributions introduced in  (\ref{eq:latentgaussian}) allow the model to condition the input into different distributions depending on class, without explicitly using the class as an input. One option is to let the mean $\mu_{y}$'s be learnable parameters of the model. This allows the model to position the clusters in a way that optimizes the performance of the reconstruction. This, however, typically leads to the model effectively covering the latent space with these clusters, leaving little room. This lack of space causes a problem when we introduce unencountered anomalous data, as the latent space now is spatially saturated with clusters. This means that anomalous points are more likely to accidentally lie in a class cluster. To combat this, we instead force the clusters away from the origin of the latent space by fixing $\mu_y$'s to points on an n-sphere. This frees up the origin and creates space for unencountered data (see Figure \ref{fig:latent_space}). The idea is that anomalies will be positioned roughly at the center of mass of the training data, due to it likely exhibiting some traits from several of the known classes. So, by positioning the clusters evenly spaced on an n-sphere, we keep the class clusters separated, while also clearing space at their center of mass, i.e. the origin of the latent space. The positioning therefore reduces to a problem of placing N points equidistantly on an n-sphere, which is an unsolved problem in general, even for the ordinary 2-sphere \cite{kogan2017new}. 

Our models will be trained on 9 classes with corresponding $\mu_y$'s. For a 2d latent space, we use equidistant points along a circle. For a 3d latent space we use the vertices of three parallel equilateral triangles, one along the equator, and two rotated $60\deg$, scaled down by $\frac{2}{3}$ and at heights $\pm \frac{\sqrt{5}}{3}$ \cite{points_on_sphere}. For an arbitrary number of classes and latent dimensions, one could employ a force simulation to converge to an optimal solution.

\subsection{Metrics and Anomaly Detection Methodology\label{sec:metrics}}

%To determine if an image is an anomaly, we first produce a latent space using the training samples. Each class cluster of the normal classes will be fitted to a simple gaussian distribution. Each image in the test case then will be encoded into the latent space, and using the class gaussians we assign a score to that data point. We used the gaussian pdf as the scoring, as it was found to be superior to other metrics such as the Mahalanobis distance. We call this metric the \textbf{latent divergence} of a data point. If the latend divergence is below a certain threshold, the image is predicted to be an anomaly. The AUC over this threshold is used as a metric of the model performance. \\
%We compare this to the traditional method where the reconstruction error is used as a proxy for how well the model understands the sample. If the reconstruction error is above a certain threshold, the image is considered an anomaly. Again, the AUC is used to compare model performances.

To assess whether an image constitutes an anomaly, a latent space is generated from the training samples. Within this space, clusters corresponding to the normal classes are each modeled with a simple Gaussian distribution. Subsequently, images from the test set are encoded into this latent space, and scores are attributed to each data point based on the class-specific Gaussians. The scoring is based on the Gaussian probability density function (pdf), which has proven more effective than alternative metrics, such as the Mahalanobis distance. We will refer to this metric as the \textbf{latent divergence} of a data point. An image is classified as an anomaly if its latent divergence falls below a predetermined threshold. The model's efficacy is gauged by the Area Under the Curve (AUC) relative to this threshold. 

This approach is contrasted with the conventional method that relies on reconstruction error as an indicator of the model's comprehension of the sample. Here, an image is deemed an anomaly if its reconstruction error surpasses a specific threshold, with the AUC metric again facilitating the comparison of model performances.

\subsection{Latent Space Ensembles}\label{sec:ensembles}

One of the benefits of VAEs is the efficient dimensionality reduction. The low dimensional data is much more suited for distance and distribution comparison because of the reduced sparsity. We do, however, lose a lot of information in the process. By utilizing the reconstruction error in the loss function, we ensure that the latent space is still suited for representing the known non-anomalous classes in the data set. However, since we do not train on the out-of-distribution data, we cannot be certain that the latent space is well equipped to represent it. The hope is that anomalous points will fall outside of the clusters of the normal classes, but with the efficient representation in the latent space, the likelihood of an anomalous point being accidentally clustered with the normal data is not insignificant.  We therefore employ an ensemble of CL-VAEs with randomized ordering of the class clusters, thus resulting in vastly different optima for the latent space. The reason for this is to increase the likelihood that two CL-VAEs will converge to different representations of the latent space. 

As discussed in Section \ref{sec:metrics}, for each model in the ensemble, a point $z$ is assigned a score for belonging to each normal class $y$ based on the density function of the latent Gaussian $p_\theta(z|y)$. To perform the ensembling, these scores are merged to an ensemble score. The arithmetic mean was found to be the superior method for ensemble merging when compared to the geometric mean, the maximum value, and the minimum value. The output of the merging also corresponds to a score for a point to belong to each class. If the largest confidence of belonging to a specific class is below a certain threshold, the point is classified as anomalous.

\subsection{Datasets and Anomaly Definition}\label{sec:datasets}

Traditional anomaly detection benchmark datasets \cite{ADBench, NumentaBench} lack class labels, precluding the possibility of conditional training. To address this, we utilize datasets such as MNIST \cite{MNIST}, Fashion-MNIST \cite{FashionMNIST}, and CIFAR-10 \cite{Cifar10}, which support conditional approaches, as evidenced in existing literature on conditional anomaly detection \cite{oskar_anomCVAE, GMMVAE}. The establishment of a standardized ground truth for anomaly detection is challenging due to the inherently subjective and task-specific nature of what constitutes an anomaly. A prevalent method involves human consensus to identify anomalies, though this approach is notably resource-intensive. An alternative, as suggested in \cite{oskar_anomCVAE}, involves deploying a classifier to infer ground truth labels based on classification loss. This, however, merely reflects the model's fidelity to the classifier, which may not be an effective anomaly detector in itself.

We adopt a more classical approach by differentiating classes within the dataset into 'normal' for training and 'anomalous' for testing purposes, effectively treating anomalies as data points absent from the training distribution.

Unlike previous models that assume a singular Gaussian distribution in latent space and train on a single class, our methodology leverages the presence of multiple class clusters. Specifically, within the MNIST dataset, digit 0 is designated as anomalous, while digits 1-9 are treated as normal. Consequently, the training dataset comprises digits 1-9 and their labels, with the test dataset containing an equal mix of normal (1-9) and anomalous (0) digits. A similar delineation is applied to Fashion-MNIST, designating class 0 (t-shirt/top) as anomalous. For CIFAR-10, class 0 (airplanes) is categorized as anomalous.

We compare the CL-VAE latent divergence results of our anomaly detection to that of the CL-VAE when evaluating on the reconstruction error, an ordinary VAE evaluating on reconstruction error, a convolutional neural network (CNN) classifier evaluating on the class confidence, and an principal component analysis (PCA) evaluating on the log-likelihood score.

For MNIST, our model architecture features a 2D latent space, two 3x3 convolutional encoding layers with a stride of 2 and 32 and 64 channels, respectively, mirrored in the decoder structure. For Fashion-MNIST, the model incorporates a 3D latent space and an additional 3x3 convolutional layer in the encoder with a stride of 1 and 16 channels. The CIFAR-10 model, tailored for its complexity, includes a 3D latent space and three 3x3 convolutional layers with strides of 2 and 64, 128, and 256 channels, respectively. This configuration is considerably simplified compared to typical models for CIFAR-10, underscoring our interest in examining whether conditioning and class separation can aid the convergence of an underdimensioned model.
We use the same layer sizes for the VAE, and a CNN size that corresponds to the same layers as the CL-VAE encoder. The full code can be found at \url{https://github.com/oskarastrom/CL-VAE/}.

\section{Anomaly Detection Results\label{sec:results}}

%Training a regular VAE on the MNIST dataset produces a latent space representation where the resulting classes accumulate around a single Gaussian. This is a direct implication of minimizing the loss function in \eqref{eq:loss_function}. We can easily visualize such a result in the top-left plots in Figure \ref{fig:latent_space} based on the MNIST dataset. The 10 MNIST handwritten numbers are all clustered around the mean of a single Gaussian and gradually expand outward from there. As can be seen in that figure, there is little class structure and heavy overlap for the VAE latent space. Class overlap is natural since handwritten numbers can sometimes resemble each other and can be hard to tell apart. These issues stem from the fact that classes are trying to adapt their form to a single Gaussian prior.

\begin{figure}[!h]
	\centering
        \includegraphics[width=1\linewidth]{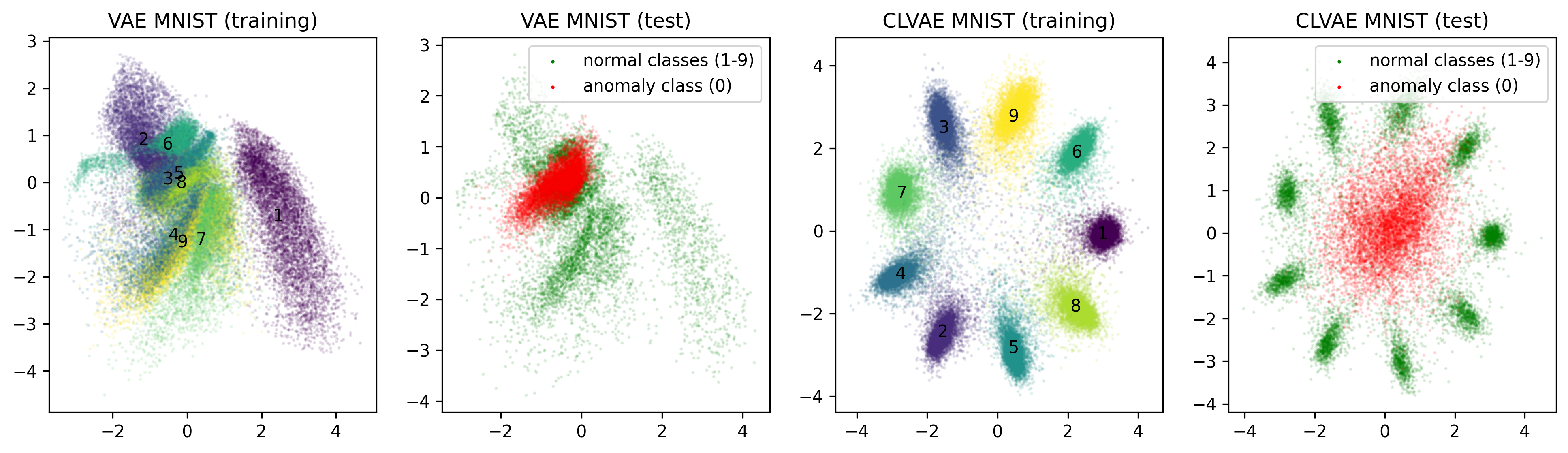}
        \hspace{-.2cm}
        \includegraphics[width=1\linewidth]{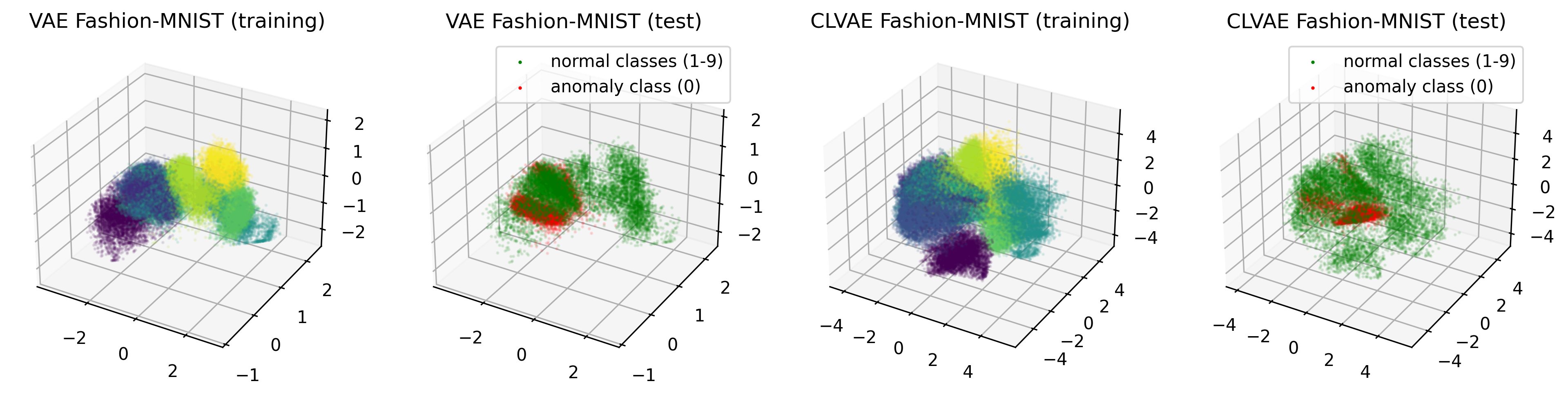}
        \hspace{-.2cm}
        \includegraphics[width=1\linewidth]{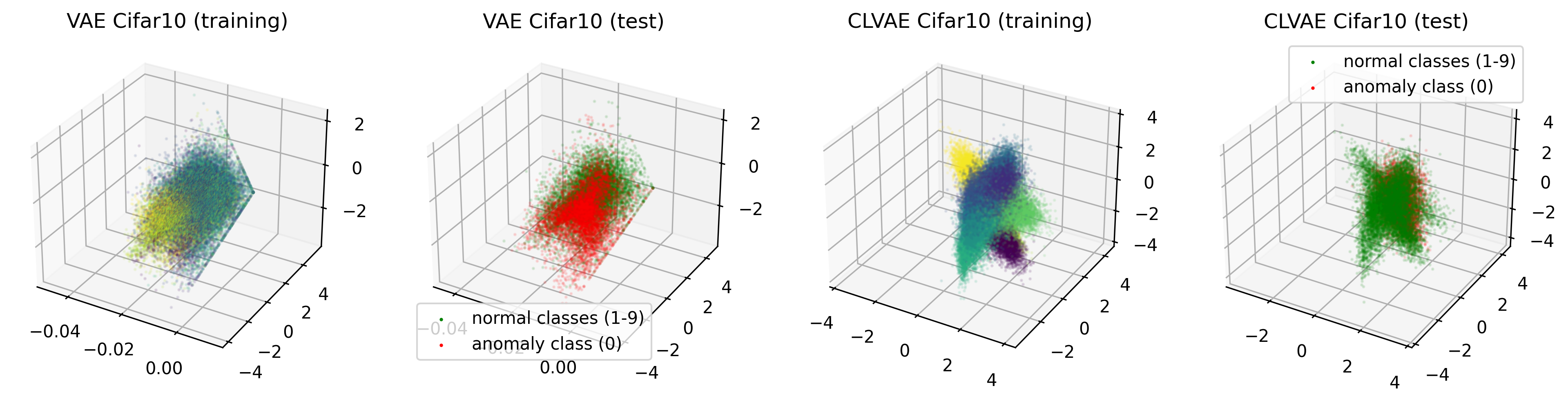}
        \hspace{-.2cm}
	\caption{Latent space representation for the three datasets (top: MNIST, center: Fashion-MNIST, bottom: CIFAR-10) using the regular VAE (two leftmost columns) and CL-VAE (two rightmost columns). For each model, two plots are presented; the training set with class centers marked with their corresponding digit, and the test set with anomalies and normal classes in different colors.}
	\label{fig:latent_space}
\end{figure}

We present the latent space after training the CL-VAE on the same data set and present the results in the top-right plots of Figure \ref{fig:latent_space}. It is immediately evident that the class separation is improved compared with that from the regular VAE. The class overlap is still visible since handwritten numbers due to their inherent shape can sometimes resemble each other in the MNIST data. In that respect it is not surprising to see that the anomalies (digit 0) are drawn to the cluster for the digit 6 in both cases. Although, this effect is much lower for the CL-VAE. Similar results are seen for the Fashion-MNIST dataset in the center row of Figure \ref{fig:latent_space}. The anomalies are overlapping with the normal classes to a larger extent than in the simpler MNIST case. The much more complicated CIFAR-10 dataset does not result in an interpretable latent space for the regular VAE, as shown in the bottom row of Figure \ref{fig:latent_space}. However, the CL-VAE still manages to clearly separate the class clusters, even with a latent space of only 3 dimensions. 

Training a standard VAE on the MNIST dataset leads to a latent space where classes tend to converge around a singular Gaussian distribution, a consequence of minimizing the loss function delineated in \eqref{eq:loss_function}. This phenomenon is depicted in the top-left plots of Figure \ref{fig:latent_space}, illustrating how the ten MNIST digits cluster around a Gaussian's mean and disperse outward. As illustrated, the VAE latent space exhibits minimal class structuring with significant overlap, a natural outcome given the visual similarities between certain handwritten digits. This overlap arises as classes conform to a singular Gaussian prior.

%Subsequently, we explore the latent space post-CL-VAE training on the same dataset, with findings showcased in the top-right plots of Figure \ref{fig:latent_space}. Visually, it becomes apparent that class distinction is notably enhanced relative to the standard VAE, despite the persistence of overlap due to the inherent resemblance among handwritten digits in the MNIST dataset. Interestingly, the anomalies (digit 0) show an affinity towards the cluster representing digit 6 in both scenarios, albeit less so in the CL-VAE.

%The Fashion-MNIST dataset, as depicted in the center row of Figure \ref{fig:latent_space}, exhibits similar tendencies, with anomalies blending more with normal classes than in MNIST. Nonetheless, the CL-VAE demonstrates a clearer class demarcation compared to the VAE.

%For the complex CIFAR-10 dataset, presented in the bottom row of Figure \ref{fig:latent_space}, the regular VAE fails to yield an intelligible latent space. Conversely, the CL-VAE achieves distinct class segregation, even within a 3-dimensional latent space.

\begin{figure}[!h]
	\centering
        \includegraphics[width=1\linewidth]{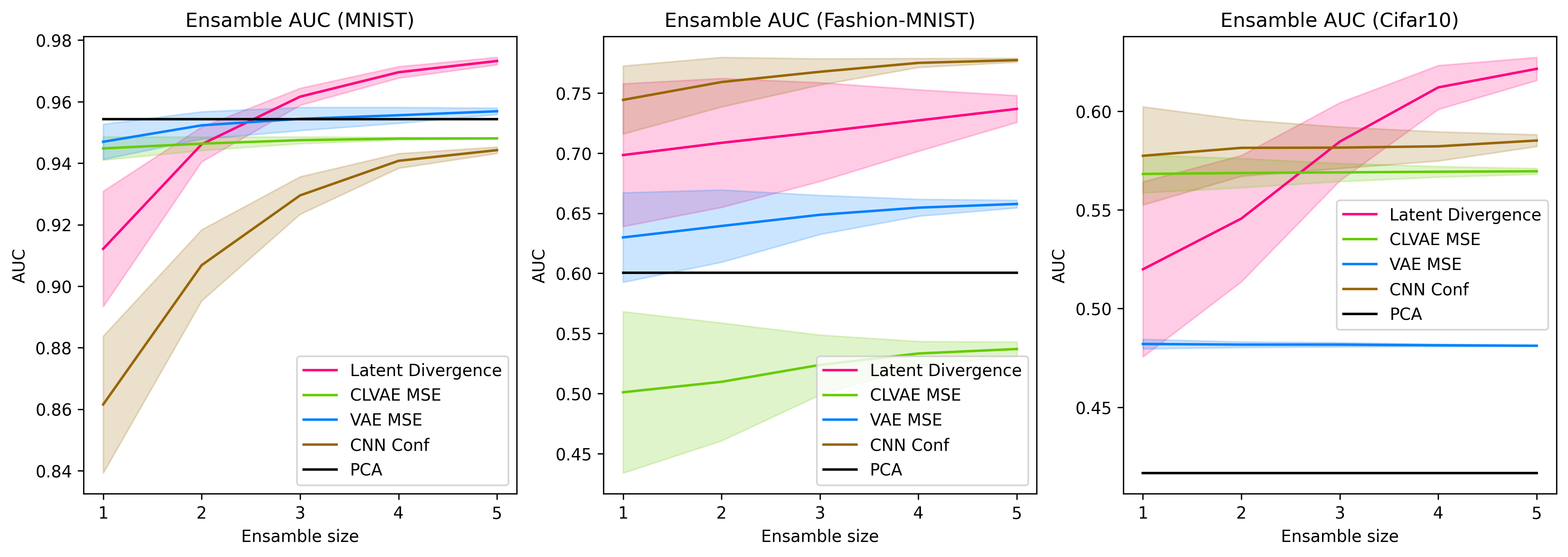}
	\caption{Confidence intervals of the AUC anomaly detection as a function of ensemble size using the 5 different evaluation methods. }
	\label{fig:ensembling_effect}
\end{figure} 

\subsection{Ensembling Effect}
Figure \ref{fig:ensembling_effect} shows confidence intervals for the AUC of the models for increasing ensemble sizes. It is evident from these figures that the CL-VAE benefits to a much larger extent from being in an ensemble compared to the VAE. For the MNIST dataset, the latent divergence overtakes the traditional VAE when using more than 2 models in the ensemble. It also outperforms both the PCA and CNN. For the more complex Fashion-MNIST and CIFAR-10 datasets, the latent divergence outperforms the VAE already at a single model. It also overtakes the CNN at 3 models in the ensemble for the CIFAR-10 dataset. For the Fashion-MNIST dataset, the CNN does however outperform the CL-VAE. This could be because the CNN has an output size of 9, whereas the CL-VAE only has a latent space of 3 dimensions. 
%As depicted in Figure \ref{fig:ensembling_effect}, the confidence intervals for the Area Under the Curve (AUC) of the models expand with the growth of ensemble sizes. These illustrations clarify that the CL-VAE experiences significantly enhanced benefits from ensembling. Specifically, within the MNIST dataset context, the performance measured by latent divergence surpasses that of the conventional VAE upon the integration of more than two models into the ensemble. In the case of the complex datasets such as Fashion-MNIST and CIFAR-10, latent divergence demonstrates superior performance to the VAE even with a solitary model in the ensemble.

\subsection{False Classifications}

\begin{figure}[!h]
	\centering
        \fbox{\includegraphics[width=1\linewidth]{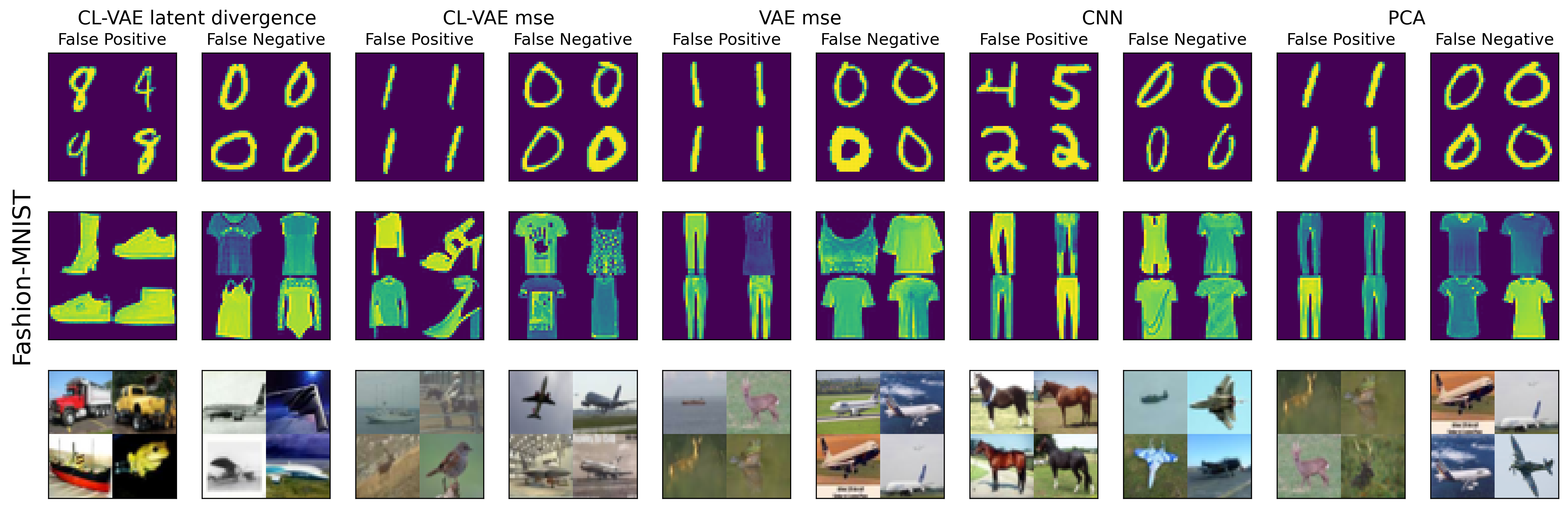}}
	\caption{Examples of misclassifications on the three datasets for each of the five evaluation methods.}
	\label{fig:misclassifications}
\end{figure}

When investigating the cases for which the models fail to identify the anomalies correctly, we find that the latent divergence captured larger variations in their misclassifications. As seen in Figure \ref{fig:misclassifications}, the latent divergence has a larger variety in their misclassifications, while the MSE metric results in a heavy reliance of the weight of the image. For example, in the MNIST dataset, the most missclassified inliers are all of 1's that don't appear too anomalous, while the missclassified outliers are all very thick 0's. The latent divergence on the other hand, results in misclassifications that do appear more anomalous, at least for MNIST. This larger variation of misclassifications could be a reason for the increased benefit of ensembling.
%During our examination of instances where the models inaccurately identified anomalies, it became apparent that latent divergence exhibited a broader spectrum of errors in its misclassifications. As illustrated in Figure \ref{fig:misclassifications}, misclassifications under latent divergence display a wider variance, contrasting with the mean squared error (MSE) metric, which tends to overly depend on the image's weight. For instance, within the MNIST dataset, the most frequently misclassified inliers predominantly include 1's, which aren't notably anomalous, whereas the incorrectly classified outliers consist chiefly of unusually thick 0's. Conversely, misclassifications associated with latent divergence seem to be more genuinely anomalous, particularly in the context of MNIST. This greater diversity in misclassifications may contribute to the enhanced effectiveness observed with ensemble methods.

\section{Conclusions and Discussion\label{discussion}}

In this article we have presented a novel method for anomaly detection based on the latent space of a variational autoencoder (VAE). Specifically we teach our autoencoder to condition on a labeled data class from which to learn an improved latent space clustering. We examine the performance of this conditioned variational autonencoder (CL-VAE) against the classic VAE, a CNN, a PCA, based on the MNIST and Fashion-MNIST datasets as well as the more complex CIFAR-10 dataset. Furthermore the proposed methodology works with data of categorical nature (non-continuous) in order to find meaningful latent representations - both of which are not possible for CNN classifier or the PCA methods \cite{PCA}.

Another advantage of the proposed CL-VAE is that it attempts to make the latent space more understandable and suitable for analysis with established methods. It seems to succeed in this regard, as it both divides the latent space up in accordance with class labels and ensures that points that are improbable do in fact end up on the tail of their respective prior distributions or in other clusters. 

We began with the MNIST data in Section \ref{sec:results} and a visual comparison of the corresponding latent spaces for the VAE and the CL-VAE which shows (see Figure \ref{fig:latent_space}) a much cleaner separation. Using these latent spaces we performed anomaly detection using the latent divergence from the class clusters and found that indeed the CL-VAE is able to identify the unseen class in the MNIST dataset. Furthermore, as seen in Figure \ref{fig:ensembling_effect}, the latent divergence realized a substantial increase in performance when used in an ensemble compared to using the other methods, thus highlighting further benefits of this method.

For the Fashion-MNIST, the latent divergence actually performed better than both MSE metrics even with only a single model. It did, however, not outperform the CNN classifier. Finally, for the much more complicated CIFAR-10 dataset, even the MSE from the CL-VAE performed better than the MSE from the standard VAE, showing that for harder problems, the CL-VAE is able to learn the latent space with limited resources. Here, the CNN outperformed the CL-VAE at 1 model, but the CL-VAE showed an massive increase in performance through ensembling, thereby overtaking the CNN at 3 models in the ensemble.

This highlights a benefit but also a drawback of this method. That is, that the CL-VAE can require an ensemble of multiple models to reach optimal performance. This of course increases training and evaluation time. This is partially mitigated by the fact that only the encoder is needed in evaluation, thus halving the storage space and evaluation time compared to using the MSE. In addition, for the more complex datasets, the CL-VAE seem to be performing better with less resources, opening up the opportunity for smaller models. This should be investigated further to see if these benefits outweigh the costs, especially on real world online tasks.
Furthermore, while the trends of increased benefit of ensembling and easier convergence for hard problems are clear, the behavior is still very different for the 3 datasets, highlighting the need for further investigation into this dynamic which could potentially help solidify the field of anomaly detection.

%This means that we currently do not have a general version of the model that can be applied in multiple areas and tasks. This is likely partly due to the fact that an "anomaly" is such a nebulous term that take different shapes for different tasks. However, further research into connections between the model parameters and tasks could potentially shed some light into this dynamic and 

\section*{Acknowledgments}
The work of O.~Å. and A.~S. is partially supported by grants from eSSENCE no. 138227,  FORMAS no. 2022-151862 and Rymdstyrelsen no. 2022-00282. The computations were enabled by resources provided by the National Academic Infrastructure for Supercomputing in Sweden (NAISS), partially funded by the Swedish Research Council through grant agreement no. 2022-06725.

\bibliographystyle{splncs04}
\bibliography{cite}

\newpage
\section*{Appendix A. Variational autoencoders and Bayes' rule}

Variational autoencoders (VAEs) are able to discover the distributions responsible for the provided data. A VAE therefore solves the problem of probability density estimation and is a true generative model. This practically means that it can generate new samples from an unknown distribution \cite{Stanford}. Applications in image processing for example use VAEs to generate new images which retain some of the main features of the original data set \cite{TrainVAE}.

%The idea is quite simple. Let's assume I want to sample from a data set $\{x_i\}_{i=1}^{N}$ that consist of N i.i.d random variables with unknown distribution. We assume that this $x$ is part of some stochastic process that involves a hidden random variable $z$ with PDF $p_{\theta^*(z)}$ that we assume is Gaussian. In this case $\theta^*$ represents the true model parameters. This process generates samples in two steps:
%\begin{enumerate}
%    \item Sample $z$ from $p_{\theta^*}(z)$
%    \item Sample $x$ from $p_{\theta^*}(x|z)$
%\end{enumerate}

If we have some set of locally observed variables $x$ and we assume that they follow some unknown stochastic process $X$ that we want to sample from, we can use some prior $z$ that we assume to be Gaussian. Then taking the expectation of the conditional distribution of $x$ given $z$ is obtained by marginalizing out the latent variable $z$,
%, under $z$, we get the distribution for $x$ from,
\begin{equation}
p_{\theta}(x) = \mathbb{E}_z\Big[p_{\theta}(x|z)\Big] = \int p_{\theta}(x|z) p_{\theta}(z) dz.
\label{eq:prob_x}
\end{equation}
Note however that the integral above is intractable \cite{VAE} as it would take exponential time to compute. This is where Bayes' rule can help. We assume that the density functions  $p(x)$ and $p(y)$ for stochastic variables x and y are known. If the conditional density function $p(x|y)$ is given then the conditional density function $p(y|x)$ can be computed from,
\begin{equation}
\label{def:bayes}
	p(y|x) = \frac{p( x| y) p(y)}{p(x)}.
\end{equation}

\noindent Bayer rule will be useful in terms of computing the posterior $p_{\theta}(z|x)$,
\begin{equation}
p_{\theta}(z|x) = \frac{p_{\theta}(x|z) p_\theta(z)}{p_\theta(x)}.
\label{eq:posterior}
\end{equation}
However $p_\theta(x)$ is not possible to compute in general as pointed out earlier. Instead we estimate the posterior distribution $p_{\theta}(z|x)$ using 
variational inference by a family of distributions $q_\phi(z|x)$. As a simple example if $q$ is Gaussian then $\phi$ is the mean and variance of each datapoint $\phi_{x_i} = (\mu_{x_i}, \sigma^2_{x_i})$.
%some other distribution  $q_\phi(z|x)$ where $\phi$ contains estimates of the model parameters. 

\newpage
\section*{Appendix B. Kullback-Leibler divergence}

The KL-divergence has a number of imporant properties which we outline below. First we provide some definitions from information theory \cite{SK}. 

Given two probability distributions $Q$ and $P$ the entropy of $P$ is defined by,
\begin{equation}
	H(P) = \mathbb{E}_{x\sim P}[-\log P(x)],
	\label{def:entropy}
\end{equation}
	and the cross-entropy of $Q$ and $P$ is given by,
\begin{equation}
	H(P, Q) = \mathbb{E}_{x\sim P}[-\log Q(x)]. 
	\label{def:cross-entropy}
\end{equation}

The KL-divergence by taking the cross-entropy minus the entropy,
\begin{align}
\label{def:KL}
\begin{split}
    D_{KL}(P||Q) = H(P, Q) - H(P) = 
    %&D_{KL}(P||Q) = 
    \mathbb{E}_{x\sim P}\Bigg[\log\frac{P(x)}{Q(x)}\Bigg].
\end{split}
\end{align}
The KL-divergence measures how well $Q$ approximates $P$, in the following sense:

\begin{property} \label{prop:KL}
	Properties of KL-divergence \cite{Tiao, der}.
	\begin{enumerate}
		\item if $P=Q$ then $D_{KL}(P||Q) = 0$,
		\item if $P\neq Q$ then $D_{KL}(P||Q) > 0$.
	\end{enumerate}
\end{property}

\begin{property} \label{prop:KL_gauss}
	Solution to $D_{KL}(q(x)||p(x))$ in the normal Gaussian case \cite{VAE}. Let's assume that x is some random variable in  $q(x)\sim N(\mu, \sigma)$ and $p(x) \sim N(0, I)$. 
	\begin{align*}
	D_{KL}(q(x)&||p(x
	)) = \mathbb{E}_{x \sim q_\phi(x)}\Bigg[\log\frac{q(x)}{p(x)}\Bigg] =\int \log \frac{q(x)}{p(x)} q(x) dx =\\
        &= \int (\log q(x) - \log p(x)) q(x) dx = \\
        &= \int q(x) \log q(x) - q(x)\log p(x)dx =\\
        &= \int N(x; \mu, \sigma^2)\log N(x; \mu, \sigma^2)d - N(x; \mu, \sigma^2) \log N(x; 0, I) dx = \\
	&= \frac{n}{2}\log(2\pi) + \frac{1}{2} \sum_{i=1}^n(1 + \log \sigma_i^2) - 
	\frac{n}{2} \log(2\pi) - \frac{1}{2} \sum_{i=1}^n (\mu_i^2 + \sigma_i^2),
	\end{align*}
	where $n$ is the number of samples in the batch. This results in the closed form solutions
	\begin{align*}
	D_{KL}(q(x)||p(x)) = 
	-\frac{1}{2}\sum_{j=1}^J 1 + \log \sigma_j^2 - \mu_j^2 - \sigma_j^2.
	\end{align*}
\end{property} 

\section*{Appendix C. Simplification of conditional KL-loss}

We consider the non-normal Gaussian case and, following ideas from \cite{one-step}, produce a closed form solution for the conditional KL loss,
%To produce a closed form solution, for the non-normal Gaussian case, the non-normal Gaussian case for the conditional KL loss,
\begin{equation}
\begin{split}
    L_{KL} = D_{KL}(q_\phi(z|x,y) || p_{\theta}(z|y))= \mathbb{E}_{q_\phi(z|x,y)}\Bigg[\log\frac{q_\phi(z|x,y)}{p_\theta(z|y)}\Bigg].
\label{eq:KL_log_2}
\end{split}{}
\end{equation}
We let $q_\phi(z|x)$ be some Gaussian distribution that is conditioned on $x$, and further, we let $p_\theta(z|y)$ be Gaussian with mean $\mu_y$ and variance 1:
 \begin{equation}
    q_\phi(z|x) = \frac{\exp\Big\{-\frac{1}{2} \Big|\Big|\frac{z-\mu(x)}{\sigma(x)}\Big|\Big|^2 \Big\}}{\prod_{i=1}^d \sqrt{2\pi\sigma_i^2(x)}}, \hspace{0.5cm} p_\theta(z|y) = \frac{1}{(2\pi)^{d/2}} \exp\Big\{-\frac{1}{2}||z-\mu_y||^2 \Big\}.
\label{eq:appendix_latentgaussian}
\end{equation}
These are used to simplify the logarithm in equation (5) in the main paper, resulting in
$$
\log\frac{q_\phi(z|x)}{p_\theta(z|y)} =  -\frac{1}{2}\sum_{i=1}^d \log\sigma_i^2(x) - \frac{1}{2} \Bigg|\Bigg|\frac{z-\mu(x)}{\sigma(x)}\Bigg|\Bigg|^2 + \frac{1}{2} ||z-\mu_y||^2.
$$
Inserting this back into the expectation from the KL-divergence gives
\begin{equation}
\begin{split}
D_{KL}(q_\phi&(z|x,y) || p_{\theta}(z|y))=\\
&=  -\frac{1}{2}\mathbb{E}_{q_\phi(z|x,y)}\Bigg[
\sum_{i=1}^d \log\sigma_i^2(x) + \Bigg|\Bigg|\frac{z-\mu(x)}{\sigma(x)}\Bigg|\Bigg|^2 - ||z-\mu_y||^2
\Bigg]   \\
&=-\frac{1}{2}\left(\sum_{i=1}^d \log\sigma_i^2(x) + \mathbb{E}_{q_\phi}\Bigg[\Bigg|\Bigg|\frac{z-\mu(x)}{\sigma(x)}\Bigg|\Bigg|^2\Bigg] - \mathbb{E}_{q_\phi}\Bigg[||z-\mu_y||^2\Bigg]\right).
\end{split}
\label{eq:KL3}
\end{equation}
The second term in the equation above is equal to $d$, as z belongs to the distribution $N(\mu(x), \sigma(x))$ under $q_\phi$. This term is the expected distance to the mean when normalized to $N(0,1)$, which is equal to the sum of $d$ variances each normalized to 1. The third term can be expressed as,
\begin{equation}
\begin{split}
\mathbb{E}_{q_\phi}\left[||z-\mu_y||^2\right] &= 
\mathbb{E}_{q_\phi}\left[(z-\mu(x)+\mu(x)-\mu_y)^T(z-\mu(x)+\mu(x)-\mu_y)\right] \\
&=\mathbb{E}_{q_\phi}\left[||z-\mu(x)||^2\right]+2\mathbb{E}_{q_\phi}\left[(z-\mu(x))^T(\mu(x)-\mu_y)\right]) + \\
&+ \mathbb{E}_{q_\phi}\left[||\mu(x)-\mu_y||^2\right] 
= 
||\sigma(x)||^2 + 0 + ||\mu(x)-\mu_y||^2. 
\end{split}
\end{equation}
Inserting this into (\ref{eq:KL3}) yields,
$$
D_{KL}(q_\phi(z|x,y) || p_{\theta}(z|y)) =
-\frac{1}{2}\left[d + \sum_{i=1}^d \log\sigma_i^2(x)\right] - ||\sigma(x)||^2 - ||\mu(x)-\mu_y||^2.
$$

\section*{Appendix D. The change of variables idea}

Functions such as (2) are not uncommon in optimization problems \cite{Helmholtz}. The novel idea, introduced in \cite{VAE}, which removes  difficulties of the expected value in 
(2) is to introduce a suitable change of variables.
%In fact most of the math up until this point was already established before the paper by Kingma and Welling in 2014. As early as 1995 the Helmholtz machine was described by Hinton et al\cite{Helmholtz} but the problem was that it is very hard to optimize the ELBO with respect to the model parameters $\phi$ and $\theta$ and led to estimates with high variance \cite{CMU}. The key insight that Kingma and Welling introduced was the reparmeterization trick which is a straightforward change of variables.

We express $z\sim q_\phi(z|x)$ through a deterministic transformation $g_\phi$ and a random variable $\epsilon \sim p(\epsilon)$ as %a function of another random variable $\epsilon$ as \cite{Tiao}, % and input $x$ with parameters $\phi$ \cite{Tiao}.
%\begin{equation*}
$
z = g_\phi(x, \epsilon),
$
%\end{equation*}
where $p(\epsilon)$ is a simple distribution which does not depend on $x$ or $\phi$. In our case we choose $p(\epsilon) = N(0,1)$ since we want the latent space to be Gaussian. This reduces $z$ to,
\begin{equation*}
z = g_\phi(x,\epsilon) = \mu_\phi(x) + \sigma_\phi(x) \odot \epsilon, \quad \epsilon \sim N(0, 1) 
\end{equation*}
We now define an auxiliary function $f(x, z) = \log p_\theta(x, z) - \log q_\phi(z|x)$ over the distribution $q_\phi(z|x)$ and write (2) as an expectation of $f$. 
%We can now write  . 
If we then substitute $z$ and compute the gradient with respect to $\phi$ we get,
\begin{align*}
\nabla_\phi \mathbb{E}_{z \sim q_\phi(z|x)}[f(x,z)] &= 
\nabla_\phi \mathbb{E}_{g_\phi(x, \epsilon) \sim p(\epsilon)}[f(x, g_\phi(x, \epsilon))] \\
&= 
\mathbb{E}_{g_\phi(x, \epsilon) \sim p(\epsilon)}[\nabla_\phi f(x, g_\phi(x, \epsilon))].
\end{align*}
The expectation and the gradient commute which allows us to practically optimize the above using back propagation.

\end{document}